\documentclass[10pt,twocolumn,letterpaper]{article}

\usepackage[pagenumbers]{cvpr} 

\usepackage{bm}
\usepackage{booktabs}
\usepackage{array,multirow,graphicx}
\usepackage{pifont}
\usepackage[usenames,dvipsnames]{xcolor}

\newif\ifdraft
\draftfalse
\drafttrue 

\definecolor{orange}{rgb}{1,0.5,0}
\definecolor{violet}{RGB}{70,0,170}
\definecolor{magenta}{RGB}{170,0,170}
\definecolor{dgreen}{RGB}{0,150,0}
\definecolor{lightskyblue}{rgb}{0.53, 0.81, 0.98}

\ifdraft
 \newcommand{\PF}[1]{{\color{orange}{\bf PF: #1}}}
 
  \newcommand{\AD}[1]{{\color{dgreen}{\bf AD: #1}}}
 
  \newcommand{\MS}[1]{{\color{blue}{\bf MS: #1}}}
 
 \newcommand{\ME}[1]{{\color{cyan}{\bf ME: #1}}}

 \newcommand{\todo}[1]{{\color{red}#1}}
 \newcommand{\TODO}[1]{\textbf{\color{red}[TODO: #1]}}

\else
 \newcommand{\PF}[1]{}
 
 \newcommand{\AD}[1]{}
  
 \newcommand{\MS}[1]{}
 
 \newcommand{\ME}[1]{}
 
  \newcommand{\todo}[1]{}
 \newcommand{\TODO}[1]{}
\fi

\newcommand{\comment}[1]{}
\newcommand{\parag}[1]{\vspace{-3mm}\paragraph{#1}}

\newcommand{\bJ}{\mathbf{J}}

\newcommand{\bX}{\mathbf{X}}

\newcommand{\bx}{\mathbf{x}}

\newcommand{\x}{\mathbf{x}}

\newcommand{\Ours}{CLOAF}

\definecolor{cvprblue}{rgb}{0.21,0.49,0.74}
\usepackage[pagebackref,breaklinks,colorlinks,citecolor=cvprblue]{hyperref}

\def\paperID{4022} 
\def\confName{CVPR}
\def\confYear{2024}

\title{CLOAF: CoLlisiOn-Aware Human Flow}

\author{
    Andrey Davydov\thanks{This work was supported in part by the Swiss National Science Foundation}
    \quad Martin Engilberge
    \quad Mathieu Salzmann
    \quad Pascal Fua\\
CVLab, EPFL\\
{\tt\small \{name\}.\{surname\}@epfl.ch}
}

\begin{document}

\maketitle

\begin{abstract}
 
Even the best current algorithms for estimating body 3D shape and pose yield results that include body self-intersections. In this paper, we present \Ours{}, which exploits the diffeomorphic nature of Ordinary Differential Equations to eliminate such self-intersections while still imposing body shape constraints. We show that, unlike earlier approaches to addressing this issue, ours completely eliminates the self-intersections without compromising the accuracy of the reconstructions. 
Being differentiable, \Ours{} can be used to fine-tune pose and shape estimation baselines to improve their overall performance and eliminate self-intersections in their predictions. Furthermore, we demonstrate how our \Ours{} strategy can be applied to practically any motion field induced by the user.
\Ours{} also makes it possible to edit motion to interact with the environment without worrying about potential collision or loss of body-shape prior.
  
\end{abstract}  

\section{Introduction}
\label{sec:intro}
\begin{figure}[t!]
 \centering
  \includegraphics[width=\linewidth]{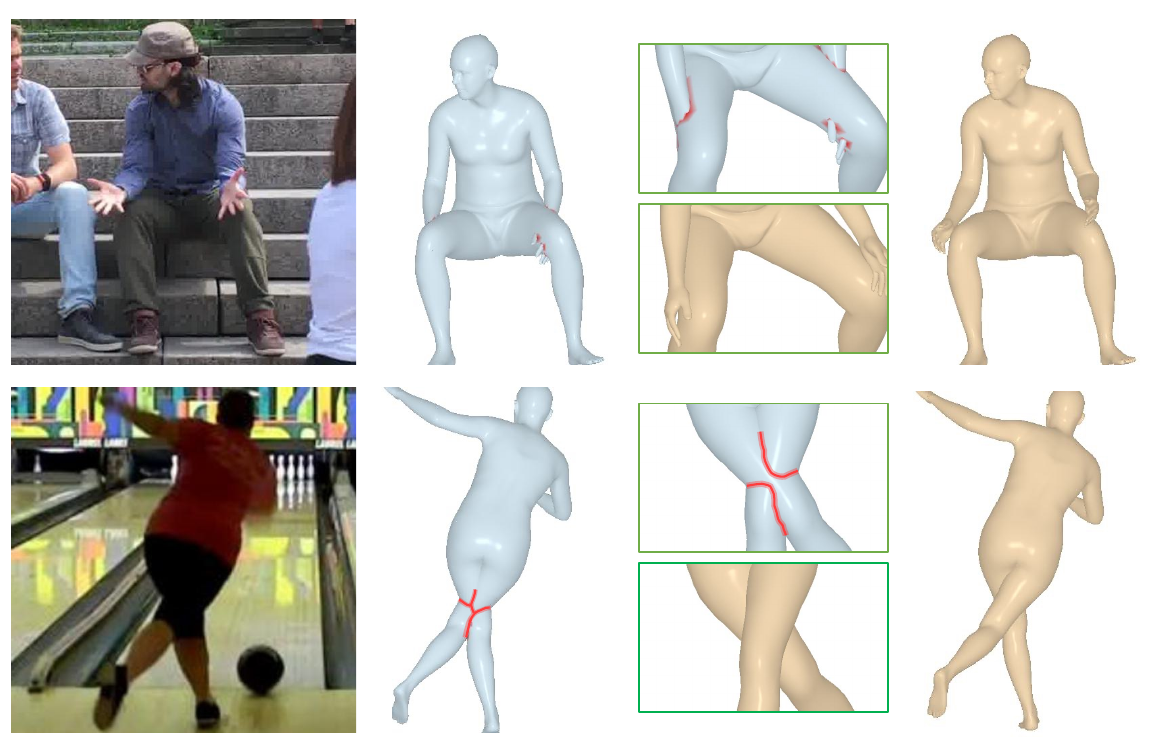} \\
  \includegraphics[width=\linewidth]{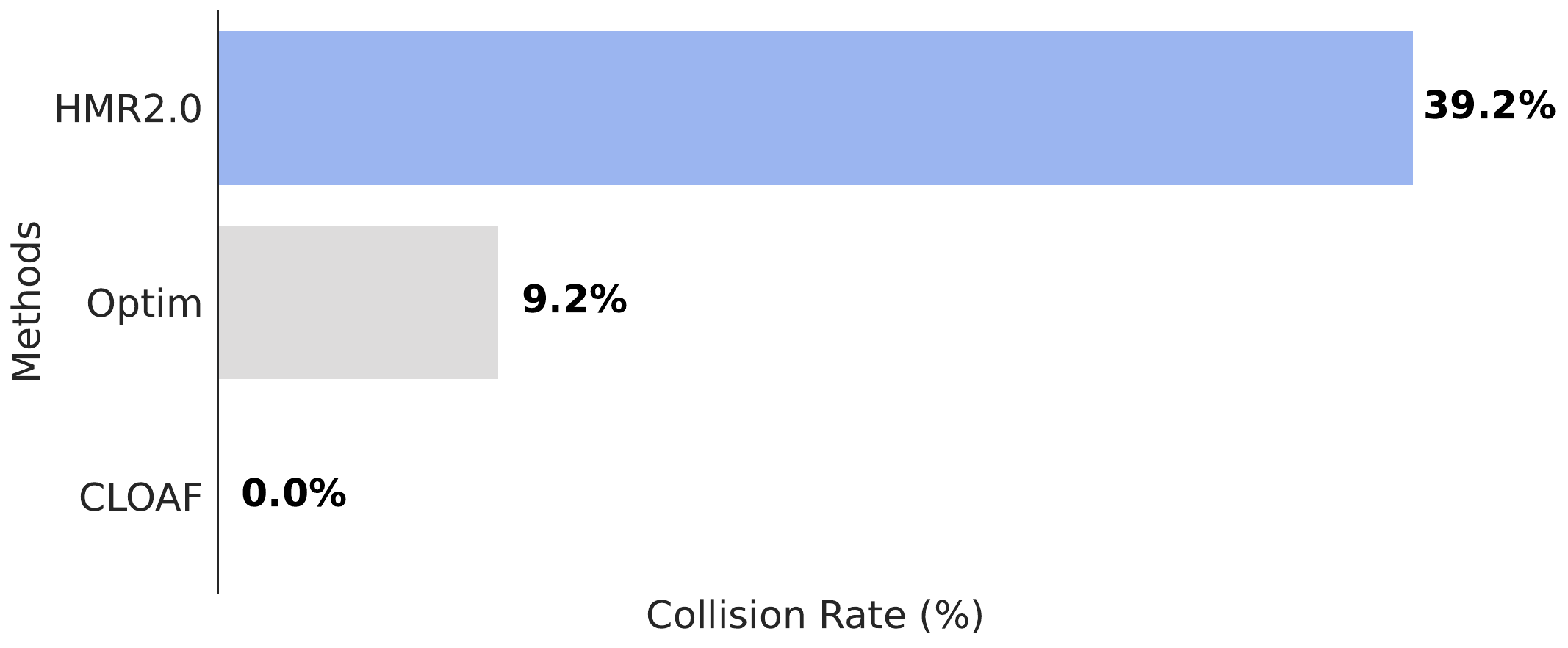}
   \caption{
       \small{{\bf Self-intersections in SOTA methods.}  
       {\it Top rows.} HMR2.0~\cite{Goel23a} ({\it first row}) and PARE~\cite{Kocabas21} ({\it second row}), two of the best current methods, produce bodies shown in {\color{RoyalBlue} blue} with self-intersections. \Ours{} removes them and generates the results shown in {\color{YellowOrange}gold}. 
       {\it Bottom row.} HMR 2.0~\cite{Goel23a} recovers bodies with self-intersections in 39.2\% of frames of the 3DPW-test set. A recent post-processing method such as~\cite{Pavlakos19a} brings this down to 9.2\%. \Ours{} drops this number all the way to zero. }
}
    \label{fig:teaser}
\end{figure}

Feed-forward approaches to estimating human body 3D shape and pose from a single image have become remarkably effective~\cite{Moon20,Choi21,Wei22}. 
The very recent transformer-based architecture of~\cite{Goel23a} embodies the current state-of-the-art. It is pre-trained on 300 million images and fine-tuned on most SMPL data sets in existence. 
However, as good as these methods have become, they can still produce unrealistic poses with substantial self-intersections of body parts, as illustrated by Fig.~\ref{fig:teaser}. This is a serious issue if video-based motion capture is to be used in fields, such as robotics or realistic animation, where preventing self-intersections is of utmost importance. 

Most current approaches to addressing this issue~\cite{Bogo16,Pavlakos19a,Hassan19,Mihajlovic22} are iterative. They penalize self-intersections explicitly by minimizing an interpenetration loss. 
This requires explicitly detecting self-intersections and then performing a separate optimization step, which makes the whole process non-differentiable and precludes its use during training.
Another approach is to eliminate self-intersections in the training databases~\cite{Mueller21}. While all these methods help, they do not guarantee the absence of self-collisions at inference time. 


In this paper, we propose a different approach. It prevents self-intersections in a differentiable manner and without an explicit detection step. 
To this end, we rely on the fact that if the scene flow from one body to another is the solution of an Ordinary Differential Equation (ODE), then there {\it cannot} be any self-intersections. 
Thus, given a volumetric representation of body shapes, we formulate an ODE that models their deformation over time and show how it can be solved with respect to the parameters of the body model we use to represent humans. This means that we can prevent self-intersections while imposing the proper geometric priors on our reconstructions.
In other words, our method is able to map any motion flow, even those that may seem implausible, to a flow without any self-intersections. None of the existing methods can achieve this.

In its simplest form, our CoLlisiOn-Aware Flow (\Ours{}) method can be used to interpolate between two non self-intersecting body representations so that the intermediate body shapes are both realistic and self-intersection free. It can also be used in a more sophisticated manner to remove self-intersections from the output of single-frame pose estimators, such as~\cite{Goel23a}, while remaining as close as possible to the original poses. Fig.~\ref{fig:teaser} illustrates this. Because it is differentiable, \Ours{} can also be integrated into the training pipeline of a deep network to improve its performance.
Additionally, we demonstrate how our \Ours{} integration procedure can utilize practically any customized motion field to move towards a target area and model interactions with surrounding objects.

In short, our contribution is to use the diffeomorphic nature of ODEs to build a flow-based pipeline to compute human body trajectories without self-intersections. This is a very generic method and, because it is differentiable, it can be used in conjunction with any body pose estimation scheme. The code will be made available at \url{https://github.com/cvlab-epfl/CLOAF}.

\section{Related Work}
\label{sec:related}

\paragraph{Feed-Forward Pose and Shape Estimation.}

While recent approaches to pose and shape estimation from images have become spectacularly good~\cite{Moon20,Choi21,Wei22,Mueller21,Kocabas21,Goel23a}, they still do not guarantee that the resulting body models are self-penetration free. 
The latest transformer-based architecture of~\cite{Goel23a} underwent pre-training using 300 million images and further refinement on the majority of existing SMPL datasets, making it the current state-of-the-art. Nevertheless, self-intersections can easily be found in its output, as shown in Fig.~\ref{fig:teaser}.  


\begin{table}[t]
    \begin{small}
    \begin{center}
    \scalebox{0.95}{
    \begin{tabular}{r||c|c|c|c}
      SOTA method & @0 & @100 & @100 {\scriptsize (w/o col.)} & P-MPJPE \\
      \hline
      HMR2.0~\cite{Goel23a} & 39.2 & 21.6 & 12.3 & 54.3 \\
      PARE~\cite{Kocabas21}   & 36.1 & 24.3 & 11.0 & 50.9 \\
      TUCH~\cite{Mueller21}  & 23.7 & 11.7 & 8.6  & 55.5 \\
      EFT~\cite{Joo20}    & 17.8 & 6.5  & 4.5  & 58.1 \\
      SPIN~\cite{Kolotouros19}   & 15.5 & 5.5  & 2.8  & 59.2 \\
        \hline
    \end{tabular}}
    \end{center}
    \end{small}
    \caption{\small {\bf Collisions in SOTA methods.} We report the Col.Rate across samples of 3DPW-test set with at least one collision (@0), at least 100 collisions (@$100$), and at least 100 collisions among samples of 3DPW-test that do not have collisions.}
    \label{tab:sota_collisions}
    \end{table}

In part, this is because most current approaches favor the accuracy of the 
body reprojection in the image, potentially at the expense of plausibility, which includes preventing self-intersection. We are not the first to notice this problem and attempts have been made to fix it. 
For example, the approach of~\cite{Mueller21} aims to make the self-contacts natural. To this end, it generates pseudo ground-truth data that features them. 
A feed-forward model trained on such data handles self-contacts better than previous methods. However, it still produces a significant number of self-intersections, as shown in Table~\ref{tab:sota_collisions}.

\parag{Collision-Aware Optimization.}
Since preventing feed-forward methods from producing self-intersections is hard, an alternative is to post-process the results to eliminate them. For example, in SMPLify~\cite{Bogo16}, limbs are modeled as ellipsoids and inter-penetrations are explicitly penalized with the corresponding loss. The more recent SMPLify-X~\cite{Pavlakos19a} uses Bounding Volume Hierarchies (BVHs) for fast collision detection and introduces local conic 3D distance fields to penalize the penetration~\cite{Ballan12,Tzionas16}. The first method is simple to use but modeling body parts as ellipsoids is an oversimplification that can yield unrealistic results. The second avoids this problem but the computation of the BVHs is costly. 
SMPLify-DC~\cite{Mueller21} extends SMPLify by modeling self-contacts more precisely. Furthermore, PROX~\cite{Hassan19} introduces inter-penetration constraints to prevent collisions between  bodies and surrounding objects. 
COAP~\cite{Mihajlovic22} suggests using independent body part-aware volumetric occupancy networks. The self-collision then can be seen as the intersection between the neighboring occupancy volumes.

In any event, none of these schemes are differentiable with respect to the input pose estimate. 
Hence, they cannot be incorporated in an end-to-end trainable pipeline. 
Additionally, they do not guarantee removal of all self-intersections because they rely on minimizing a loss. 
By contrast, our flow-based approach never produces self-intersections, is fully differentiable, and can be used during training.

\parag{Motion Field Integration.}

Central to our work is the idea of integrating an ODE-based field to prevent self-intersections. 

This concept has been explored since well before the deep-learning era, especially for shape transfer purposes. In \cite{Funck06}, shape deformations are modeled as local path line integrations. This enables volume-preserving transformations between shapes while avoiding self-intersections for practically any given input transformation. To preserve diffeomorphisms, NMF~\cite{Gupta20} employs a series of learnable ODE integrations to morph a spherical mesh into various shapes, conditioned by a point cloud. It is shown that the generated meshed objects retain feasible physical properties.
MeshODE~\cite{Huang20d} and ShapeFlow~\cite{Jiang20e} directly learn the volumetric field between pairs of meshed objects, then transferring one shape to another without collisions. ODE integration is also discussed in human motion modeling. OccFLow~\cite{Niemeyer19b} models the temporal deformation sequence for a single human subject using a flow field.

However, none of these methods exploits any data-driven prior model. Hence, intermediate shapes are not guaranteed to be realistic. In essence, we extend these concepts to the problem of transitioning between human body shapes while maintaining a valid parametric representation that preserves realism at every step of the integration.

\parag{Inverse Kinematics.}

One of our main contributions is the coupling of ODEs with a parametric model. The recovery of the underlying pose from spatial points resembles an inverse kinematics (IK) problem. 
HybriK~\cite{Li21i} incorporates an iterative IK module into the image-to-mesh recovery network to better align the 3D keypoints with a parametric body representation. 
Even though IK is solvable with iterative techniques, it is usually highly restricted by the spatial rig and the structure of the kinematic chains~\cite{Aristidou18}. 
Instead of performing exact but sometimes overly rigid IK,~\cite{Ciccone19} proposes to explicitly optimize the tangent vector using first-order approximations of the input field.
In \Ours, we project the input motion onto the most plausible velocity of the parametric state. Since we work with velocities instead of displacements, our inverse projection is exact and does not require iterating. 
In simple terms, we take the best of both worlds: integrating the motion flow without penetrations, while preserving a parametric representation of the body at every moment.

\section{Method}
\label{sec:method}


\begin{figure*}[t!]
    \begin{center}
    \includegraphics[width=\linewidth]{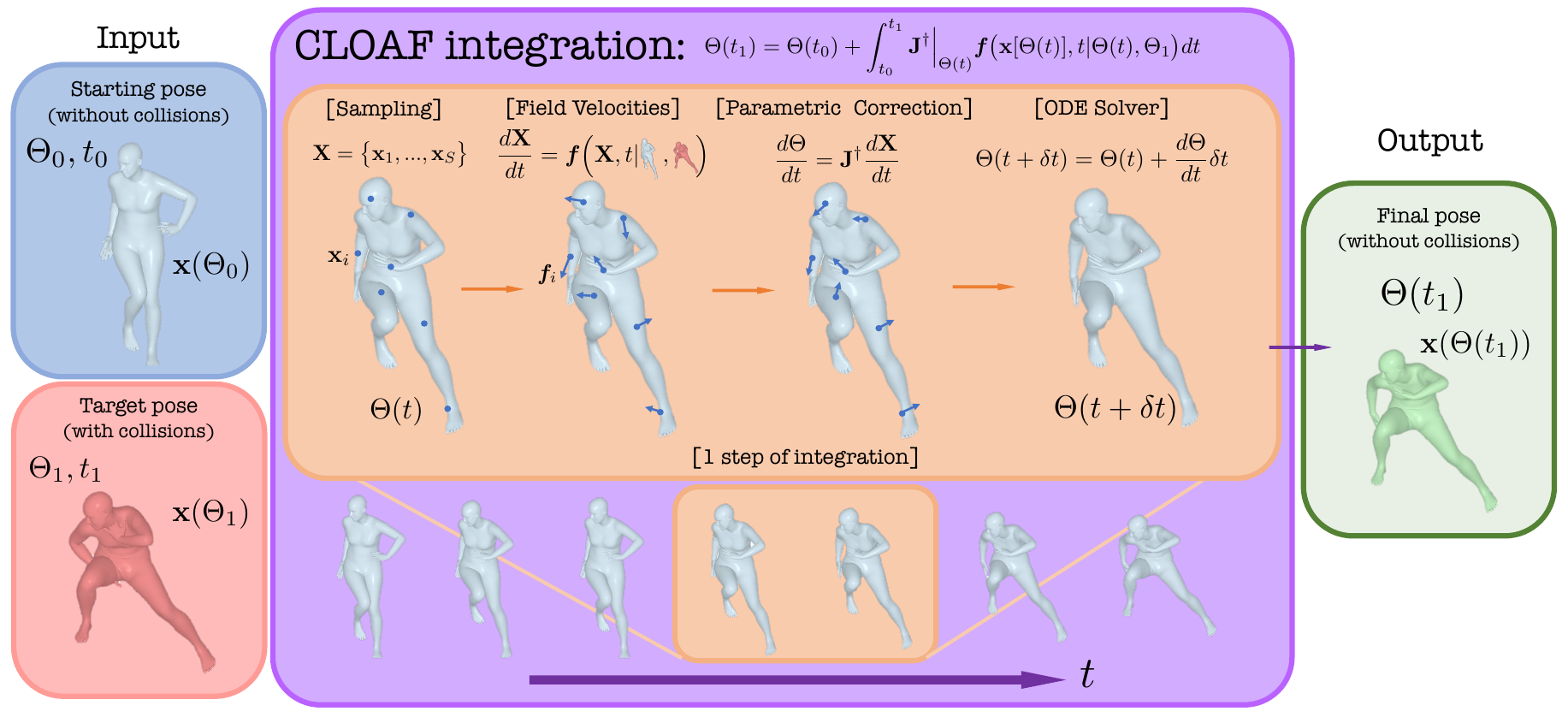}
    \end{center}
    \vspace{-2mm}
    \caption{
       \small{{\bf Method overview.} \Ours{} {\color{violet} integrates} from an  {\color{RoyalBlue} initial body pose} without self-intersections towards  a {\color{red} target one} that may feature some. 
       {\color{orange} Every integration step} involves sampling points from the body surface, calculating approximate spatial velocities, correcting these velocities in the parametric space, and performing an integration step using the ODE solver. When the integration is complete,  the body shape is without self-intersections and its pose is taken to be the  {\color{ForestGreen}  corrected pose} and the output of our method}.
       }
       \vspace{-2mm}
    
    \label{fig:overview}
\end{figure*}

When predicting body shape and pose in terms of a parameterized body model such as SMPL~\cite{Loper15}, the simplest way to discourage self-intersections is to introduce loss functions that make them costly~\cite{Bogo16,Pavlakos19a}. 
This can be effective but suffers the same fate as all soft constraints: they can still be violated. 
In this work, we exploit diffeomorphism, a key property of Ordinary Differential Equations (ODEs), to truly prevent self-intersections.

\parag{Motion Flow as an ODE.}

Let us consider a volume containing a body deforming from a first position $B_0$ to a second one  $B_1$ between times $t_0$ and $t_1$. 
Each point in that volume follows a specific trajectory. 
Let us assume the existence of a function  $\bm{f}_{\omega}$ such that we can write for every point $\x$ in the volume: 
\begin{align}
&\frac{d\x}{dt} = \bm{f}_{\omega}(\x, t | B_0, B_1) \quad \forall t, t_0 \leq t \leq t_1 \; , \label{eq:naiveOde}  \\
&\x(t_0) = \x_0 \; , \nonumber 
\end{align}
where $\x_0$ the initial position of $\x$, while the parameters $\omega$ control the behavior of $\bm{f}$.
Then, according to the Picard-Lindel{\"o}f theorem~\cite{Coddington}, the trajectories of two initially distinct points can {\it never} intersect. This requires the right-hand side of the ODE, the velocity field $\bm{f}$, to be Lipschitz continuous with respect to $\x$, which neural networks satisfy~\cite{Smith21a,Virmaux18}.

Given the formulation of Eq.~\ref{eq:naiveOde}, a start position $B_0$, and an end position $B_1$, computing trajectories for points in the volume is a classic Cauchy problem that can be solved efficiently using numerical solvers. This means that if we start from a valid position $B_0$, it will remain valid throughout $t_0 < t < t_1$. In the remainder of this section and the experiment section, we show that the function $\bm{f}_{\omega}$ exists and can be learned from data.

\parag{Introducing a Body Model.}

The function $\bm{f}_{\omega}$ of Eq.~\ref{eq:naiveOde} could be implemented by a neural network with weights $\omega$. Then, given $N$ pairs of start and end positions $\{ (B_0^i,B_1^i)  , 1 \leq i \leq N \}$ , the weights could be learned  so that the final body positions obtained by solving the ODE starting from $B^i_0$ are as close as possible to $B^i_1$.
This amounts to minimizing 
\begin{equation}
    \sum_i \| B_{0 \rightarrow 1}^i - \- B_1^i \|^2 \; ,
    \label{eq:optimLoss1}
\end{equation}
with respect to $\omega$, where $B_{0 \rightarrow 1}^i$ is the body position in the coordinate space estimated at time $t_1$, starting from $B_0$ at $t_0$. Hereafter, we omit the explicit dependency of the results of the integration on $\omega$ for notational simplicity.

However, without any body shape prior, the intermediate body positions would be completely unrealistic~\cite{Jiang20e,Niemeyer19b}. 
Thus, we propose to incorporate a body model, specifically the SMPL model~\cite{Loper15}, in this formalism. 
To this end, we reformulate Eq.~\ref{eq:naiveOde} in terms of the parameters $\Theta$ of the body model, rather than the points $\x$ as follows.

When using the SMPL model, each 3D point $\bx$ on the body surface is parameterized by the underlying SMPL vector $\Theta \in \mathbb{R}^d$. For each one, we can write
\begin{align}
    \x(t) & = \x(\Theta(t)) \; ,  \nonumber \\
   \Rightarrow \frac{d\x}{dt} & =  \frac{d\x}{d\Theta}\frac{d\Theta}{dt} = \bJ  \frac{d\Theta}{dt} \; ,  \label{eq:x_smpl}
\end{align}
where $\bJ \in \mathbb{R}^{3 \times d}$ is the Jacobian of the SMPL transformation $\x(\Theta)$, computed given the body state $\Theta(t)$. 
Injecting this into the time derivative of Eq.~\ref{eq:naiveOde} yields
\begin{align}
    \bJ \frac{d\Theta}{dt}= \bm{f}_{\omega}(\x, t | \Theta_0, \Theta_1) \; . \label{eq:lin_sys}
\end{align}
Since the number of points $S$ from the SMPL mesh can be taken to be much larger than the dimension of $\Theta$,
writing Eq.~\ref{eq:lin_sys} for each point produces an over-constrained system of linear equations. 
It can be solved in the least-squares sense, which yields
\begin{align}
    \frac{d\Theta}{dt} = \bJ^{\dagger} \cdot \bm{f}_{\omega}(\bX, t | \Theta_0, \Theta_1)  \; ,
    \label{eq:lsq}
\end{align}
where $\bX \in  \mathbb{R}^{3S}$ is the vector formed by concatenating the coordinates of all $S$ points and $\bJ^{\dagger} = (\bJ^T \bJ)^{-1}\bJ^T$ is the Moore-Penrose pseudo-inverse of $\bJ \in \mathbb{R}^{3S \times d}$. 
This computation is akin to solving an inverse kinematic problem~\cite{Grochow04,Baerloch04,Ciccone19}. 

We can now reformulate the ODE of Eq.~\ref{eq:naiveOde} in terms of the parameters $\Theta$ of the SMPL model using Eq.~\ref{eq:x_smpl}. 
It becomes
\begin{align}
        &\frac{d\Theta}{dt} =  \bJ^{\dagger}  \bm{f}_{\omega} (\bX(\Theta), t | \Theta_0, \Theta_1) \; ,   \label{eq:smplOde} \\
        &\Theta(t_0) = \Theta_0 \; , \nonumber 
\end{align}
where $\Theta_0$ parameterizes the initial body pose, $\Theta_1$ the final one, and $f_{\omega}$ approximates $\frac{d\bX}{dt}$. Solving Eq.~\ref{eq:smplOde} yields an evolution in the SMPL parameter space instead of the coordinate space of Eq.~\ref{eq:naiveOde}. Thus, it enforces the learned shape prior.

To train the network, we learn its weights $\omega$ by adapting the training scheme introduced above as follows. 
Given $N$ pairs of start and end poses $\{ (\Theta_0^i,\Theta_1^i)  , 1 \leq i \leq N \}$, the weights are learned so that the final body positions corresponding to the parameters obtained by solving the ODE starting from the initial $\Theta_0^i$ are as close as possible to the body positions corresponding to the target parameters $\Theta_1^i$. Hence, we reformulate Eq,~\ref{eq:optimLoss1} in terms of $\Theta$, which induces body points $\x$ using Eq.~\ref{eq:x_smpl}. This amounts to minimizing 
\begin{equation}
    \sum_i \| \x(\Theta_{0 \rightarrow 1}^i) - \x(\Theta_1^i) \|^2 \; ,
    \label{eq:smplLoss}
\end{equation}
with respect to $\omega$, where $\Theta_{0 \rightarrow 1}^i$ is the body position in the parametric space estimated at time $t_1$, starting from $\Theta_0$ at $t_0$. 

\parag{Points Sampling.}
Solving Eq.~\ref{eq:smplOde} requires multiple computations of the Jacobian, which means performing both the forward and backward pass of the SMPL transformation.
On modern GPUs, this can be efficiently parallelized. In practice, we found that $S=1000$ points is sufficient to cover the full body shape and compute meaningful Jacobians. We ablate the sampling number $S$ in Sec.~\ref{sec:sampling}.

In~\cite{Mueller21}, it was pointed out that some regions of the body, such as the crotch or the armpits, are more prone to natural self-intersections. Furthermore, it was shown that the SMPL blending parameters do not adequately compensate for that, being insufficiently precise~\cite{Bogo16,Pavlakos19a}. Hence, we exclude these areas from the sampling procedure.

\parag{Network Architecture and Training.}
The input vector to the motion field network $\bm{f}_{\omega}$ consists of the points $\x$ and the trajectory description $\{t, t_0, t_1, \Theta_0, \Theta_1\}$. Specific input strategies are discussed in the following paragraph. All input values are extended with Fourier features~\cite{Tancik20} to better accommodate to slight variations in the input signal, the number of frequencies is $n_f=20$.
Regarding the model, we use a variant of the ShapeFlow network architecture~\cite{Jiang20e} to implement $\bm{f}_{\omega}$. It relies on a 6-layer MLP, where the output of each layer (except the last) is concatenated with an input vector.
Note that such architecture is Lipschitz continuous, as required by the theory, because all the constituent layers are Lipschitz.

To solve ODEs, we use NeuralODE~\cite{Chen18g} with the adaptive Dormand-Prince solver. For all computations in this paper, we use one Tensor Core GPU NVidia A100.

As our training data, we use the AMASS dataset~\cite{Mahmood19} that contains pose sequences stored in the SMPL format. 
During training, we sample pairs of poses and integrate between them.
In our experiments, we found that using the absolute time $t$ as input does not provide any sufficient information to the model, preventing robust convergence. Instead, we use the ``time left'' variable, $\Delta t = t_1 - t$; this is crucial to give the model a sense of speed, and it significantly improves convergence. 

At each step, the field $\bm{f}_{\omega}$ is computed at $S$ points $\x$ from the body surface, using Eq.~\ref{eq:x_smpl}. Following~\cite{Zhang21e}, points are sampled uniformly from the mesh surface.
Finally, the motion field network can be written as
\begin{equation}
    \bm{f}_{\omega} = \bm{f}_{\omega}(\x | \Theta(t), \Theta_1, \Delta t) \; ,
    \label{eq:field}
\end{equation}
where all points $\x$ are stacked together and concatenated with the current pose $\Theta(t)$, the target pose $\Theta_1$ and the time gap $\Delta t$ that are the same for a given body. 

In our experiments, we found that the flow model trained from scratch is prone to produce unrealistic velocity values. To avoid this, we scale the output predictions by the expected speed averaged across all the points, we see that it substantially stabilizes the training and converges faster to plausible fields. 

Solving Eq.~\ref{eq:smplOde} is a  two-step operation: field estimation followed by parametric re-projection, with the latter taking time. We found that replacing $\Theta(t)$ by its approximation computed via linear interpolation 
$\tilde{\Theta}(t) = \Theta_0 + \frac{t-t_0}{t_1 - t_0} (\Theta_1 - \Theta_0)$
significantly stabilizes the training, makes it much faster, and does not bring any detrimental effect at inference time. This means that the model implicitly learns a linear interpolation in the parametric space through integration in the coordinate space, where non self-intersection is preserved.

When integrating the field for one pair of poses we induce the loss only for the ending point of the trajectory, $\Theta_1$, as shown in Eq.~\ref{eq:smplLoss}. We found that the model is prone to get stuck and not move towards the target, especially in the first stages of training. To address this, we extend the aforementioned approximation and compute the following ``trajectory'' loss for every $i$th pair of poses:
\begin{equation}
    L^i_{traj} = \frac{1}{M}\sum^M_{m=0}  \| \Theta_{0 \rightarrow t_m} - \x(\tilde{\Theta}(t_m)) \|^2 \; ,
    \label{eq:trajLoss}
\end{equation}
where $t_m = t_0 + \frac{m}{M} (t_1 - t_0)$, $M$ is the number of steps in the integration, and $\Theta_{0 \rightarrow t_m}$ is the body position in the parametric space estimated at time $t_m$, starting from $\Theta_0$ at $t_0$. Such loss significantly helps the model to stay on the trajectory and not diverge from the target.


\section{Experiments}
\label{sec:experiments}

\subsection{Datasets and Metrics}
\label{sec:datasets_and_metrics}

\parag{Datasets.}
To train the motion field network, we utilize the AMASS dataset~\cite{Mahmood19}, which contains more than 40 hours of motion sequences in the common SMPL body representation. To evaluate the baseline methods and our approach, we use the 3DPW~\cite{Marcard18} dataset. It contains 60 video sequences of various activities in the wild. All samples have a ground-truth in the SMPL format as well. We use the common train/test split, the training part is used for unsupervised fine-tuning (Sec.~\ref{sec:finetuning}), while the test set is a common benchmark and is used in all our experiments for evaluation. The $\text{COCO}_{\text{EFT}}$~\cite{Joo20} pseudo ground-truth dataset of in-the-wild images is used for supervised part of the EFT baseline training in Sec.~\ref{sec:finetuning}.

\parag{Metrics.}
To evaluate the 3D pose, we use the standard 3D Mean-Per-Joint Position Error (MPJPE, {\it mm}) and its Procrustes Aligned version (P-MPJPE, {\it mm}). 
To assess the smoothness of the motion, we compute the acceleration error (Accel.Err, $mm/s^2$) between the predicted and ground-truth 3D keypoints. This metric has been used in previous works~\cite{Kocabas20,Kanazawa19a,Choi21} to quantify the trade-off between 3D pose accuracy and motion consistency.
Since our main goal is to eliminate self-intersections, we also report the collision rate (Col.Rate@$C$, \%) computed on the 3DPW-test~\cite{Marcard18} set. It reflects the proportion of samples across the dataset that have more than $C$ vertices inside the body.

\subsection{Eliminating Self-Intersections}
\label{sec:eliminate}

As shown in Table~\ref{tab:sota_collisions}, self-collisions are prevalent in the output of some of the best current techniques. \Ours{} can be used to remove them. 


\begin{table}[t]
    \begin{small}
    \begin{center}
    \scalebox{0.89}{
    \begin{tabular}{c|c|c|c|c}
      method & MPJPE & P-MPJPE & Accel.Err & Col.Rate@0\\
      \hline
      HMR2.0~\cite{Goel23a}  & 82.0 & 52.7 & 16.1 & 39.2\% \\
      \hline
      Opt.Cones~\cite{Pavlakos19a} & {\bf 81.3} & {\bf 52.4} & 14.3 & 9.2\%\\
      Opt.Cones$^+$~\cite{Pavlakos19a} & 82.5 & 53.3 & 16.4 & 9.1\%\\
      Opt.Contact~\cite{Mueller21} & 81.7 & 53.1 & 14.9 & 8.6\%\\
      COAP~\cite{Mihajlovic22} & 81.9 & 52.8 & 14.5 & 5.7\%\\
      \hline
      \Ours & 82.3 & 53.2 & {\bf 9.4} & {\bf 0.0\%}\\
      \hline
    \end{tabular}}
    \end{center}
    \end{small}
    \vspace{-5mm}
    \caption{\small {\bf Comparison against collision penalizing techniques.}
    Our method yields smoother motion flow compared to previous approaches and guarantees collision-free predictions. In terms of pose estimation metrics, it is on par with existing methods.
    The ``$^+$'' symbol denotes that the method uses previous frames for  initialization, as we do. We use 3DPW-test~\cite{Marcard18} for evaluation.
    }
    \label{tab:col_penalizer}
    \end{table}

Let $\Theta$ be the body shape estimate produced by a neural network, which can contain self-intersections such as those depicted by Fig.~\ref{fig:teaser}. To eliminate them in any given frame of a video sequence, we start from a body shape estimate $\Theta_0$ that does not contain any and solve the ODE of Eq.~\ref{eq:smplOde} with the target shape $\Theta_1$ being $\Theta$. Body shapes along the resulting trajectory are guaranteed to be self-intersection-free and we take the final one to be our refined estimate.
Note that for this to work properly $\Theta_0$ must be self-intersection-free. As we work with complete video sequences, when processing a frame, we take the corrected pose in the previous one to be $\Theta_0$. In the case when we do not have access to sequences, we can use different strategies that we discuss in Section~\ref{sec:pick}.

In Table~\ref{tab:col_penalizer}, we compare our results against those of the baselines. For fairness, we test the baseline of~\cite{Pavlakos19a} initialized using both, the original approach (the body estimate from the current frame) and ours (the body estimated from the neighboring frame). 
While the optimization-based methods and their soft constraints cannot completely eliminate the self-intersections, ours does while at the same time providing poses that are temporally more consistent. This is achieved at the cost of a very slight drop in reconstruction accuracy, which is not clearly significant because there are some self-intersections in the so-called ground-truth data. Fixing such errors, while actually correct, makes our results appear to be further from the ground-truth than some incorrect ones.

\subsection{Self-Intersection-Aware Fine-Tuning}
\label{sec:finetuning}

\Ours{} is differentiable with respect to the corrupted input poses. Therefore, it can be added at the end of any pose and shape estimation network to fine-tune it in an end-to-end manner to reduce its self-intersection rates.


\begin{table}[t]
    \begin{small}
    \begin{center}
    \scalebox{0.9}{
    \begin{tabular}{l|cc}
      method & P-MPJPE & Col.Rate\\
      \hline
        EFT~\cite{Joo20} (baseline)  &  58.1 & 6.5 \\
        +\Ours~(post-proc.) & 57.9 & 0.0 \\
        \hline
        Opt.Cones~\cite{Pavlakos19a} & 57.5 & 6.1 \\
        \Ours~({\it no diff}) & 57.8 & 6.3 \\
        \Ours~({\it diff}) & {\bf 55.4} & {\bf 3.4} \\
    \hline
    \end{tabular}}
    \end{center}
    \end{small}
    \vspace{-5mm}
    \caption{\small {\bf Using the differentiability of \Ours{} to fine-tune a network.} This is less efficient to remove self-intersections than using \Ours{} to post-process but better than the comparable baselines.
    Fine-tuning with differentiable \Ours{} significantly improves the accuracy of the model. We use the 3DPW-test set~\cite{Marcard18} for evaluation.
    }
    \vspace{-2mm}
    \label{table:col_tuning}
    \end{table}

We demonstrate this using the EFT pose and shape estimation model \cite{Joo20} parameterized by $\Omega$. Its predictions $\hat{\x}_\Omega$ can have self-intersections. We can remove them by computing $\Ours(\hat{\x}_\Omega)$. The differentiability of \Ours{} allows us to introduce the loss $L_{\rm{cloaf}}= \|\hat{\x}_\Omega - \Ours(\hat{\x}_\Omega)\|$, whose computation does not require any new annotations. We then use the 3DPW training set~\cite{Marcard18} to refine the network weights by minimizing a composite loss that is the sum of $L_{\rm{cloaf}}$ and the usual supervised loss.

In Table~\ref{table:col_tuning}, we compare our approach to two baselines. The first is designed for comparison against a traditional optimization method~\cite{Pavlakos19a}. In this scenario, the network is fine-tuned using the loss $L = \|\hat{\x}_\Omega - \tilde{\x}\|$, where $\tilde{\x}$ is the self-intersection-free result of the optimization method. Since it is not differentiable\footnote{It should be noted that here we refer to the approach of~\cite{Pavlakos19a}, which is based on iterative optimization. Its final output is not differentiable with respect to the initial pose, while \Ours{}'s estimate is.}, the gradient of the collision correcting operation cannot be used, which hurts performance. The second baseline involves \Ours{} but with the gradient detached when computing $L$, meaning that the right term of the loss is only used in the computation of the loss, but no gradient from the ODE solver is used.

All three fine-tuning strategies help reduce the self-intersection rate, but our approach (\Ours~({\it diff})) significantly outperforms the optimization-based method in both self-intersection rate and P-MPJPE. Our ablated method with the detached gradient (\Ours~({\it no diff})) is comparable to the optimization-based method. This confirms that the benefit of our approach lies in its differentiability. 

\subsection{Simplified Motion Fields}
\label{sec:custom_field}


\begin{figure*}[t!]
    \begin{center}
    \includegraphics[width=0.95\linewidth]{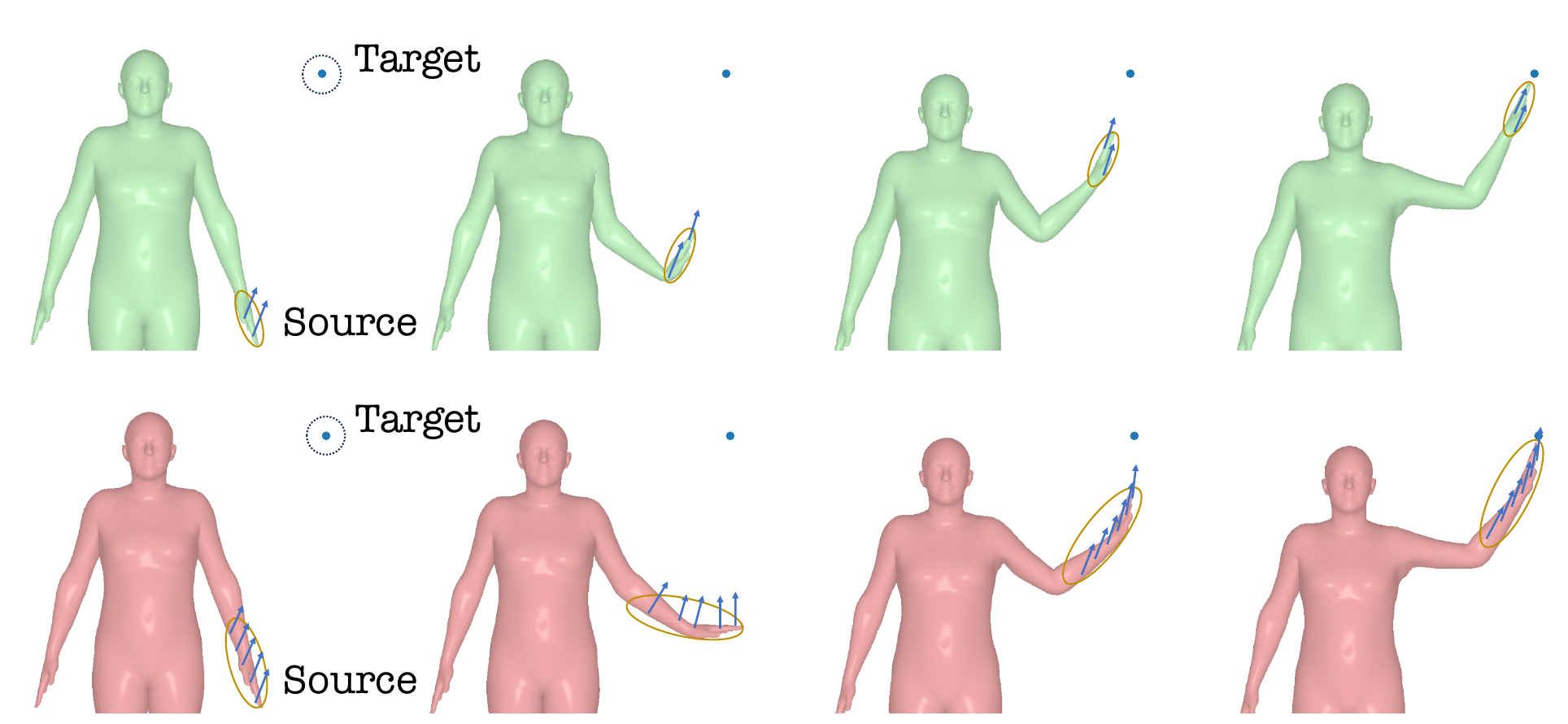}
    \end{center}
    \vspace{-3mm}
    \caption{
       \small{{\bf Moving towards a target.} 
       \Ours{} can be used to integrate practically any field, even those that are induced locally. Here the motion is set by the source region on the body (gold oval around the left arm) and the target point (blue dot).
       The field comprises the direction towards the target point (blue arrows) and adapts during the integration.  In the top row, the source region is smaller than in the bottow row, which  affects the behavior of the field and induces different motion.
       }}
       \vspace{-5mm}
    \label{fig:target_field}
\end{figure*}

So far, we have trained the motion field network $\bm{f}_{\omega}$ of Eq.~\ref{eq:field} to produce realistic motion fields and have constrained the trajectories to go from a start pose towards a final one. However, 
there are scenarios in which someone might wish to use simpler fields described by rough displacements that are defined only locally. A simple example is a piecewise field that ``moves the left arm up''. It can be defined by a vector $\bm{f}(\x) \neq 0$ around the left arm and zero elsewhere, as illustrated by the  gold ovals in Fig.~\ref{fig:target_field}. Such a field does not satisfy the Picard-Linde\"{o}f theorem, which we relied in the derivation of Eqs.~\ref{eq:naiveOde} and~\ref{eq:smplOde}. Nevertheless, it can be approximated by one that does. 
Specifically, as in~\cite{Funck06}, we blend the inner and outer regions in a Lipschitz-continuous way. 
To this end, we use a Bézier blending function 
\begin{equation}
    b(r) = \sum_{p=0}^4 \alpha_p B^4_p\Big(\frac{r - r_{\text{in}}}{r_{\text{out}}-r_{\text{in}}}\Big) \; ,
    \label{eq:blending}
\end{equation}
where $B^4_p$ are the Bernstein polynomials~\cite{Farin92}, $r_{\text{in}}$ and $r_{\text{out}}$ are the thresholds for the inner and outer regions, respectively, and $\alpha_p$ are the blending coefficients, $\alpha_0=\alpha_1=\alpha_2 = 0$ and $\alpha_3=\alpha_4=1$. 
Then, the blended field is split into three regions as follows:
\begin{equation}
    \bm{f}_b = 
    \begin{cases}
        \bm{f}(\x)  & r(\x) < r_{\text{in}} \\
        \bm{f}(\x)\cdot(1-b) + \mathbf{0}\cdot b & r_{\text{in}} \leq r(\x) \leq r_{\text{out}} \\ 
        \mathbf{0} & r(\x) > r_{\text{out}} 
      \end{cases}
     \label{eq:blended_field}
\end{equation}
This blended field  can be handled by \Ours{}. We follow the same procedure as in Sec.~\ref{sec:method}, but instead of using the neural network to estimate the motion field, we use $\bm{f}_b$.

We provide two examples in Fig.~\ref{fig:target_field}. In both cases, we start with the same body posture, and the goal is to move the left arm towards the blue dot, but we change the size of the non-zero component of the field (gold ovals). The smaller the region, the less realistic the motion is, since fewer body parameters are affected. 


\begin{figure}[t!]
 \centering
  \includegraphics[width=\linewidth]{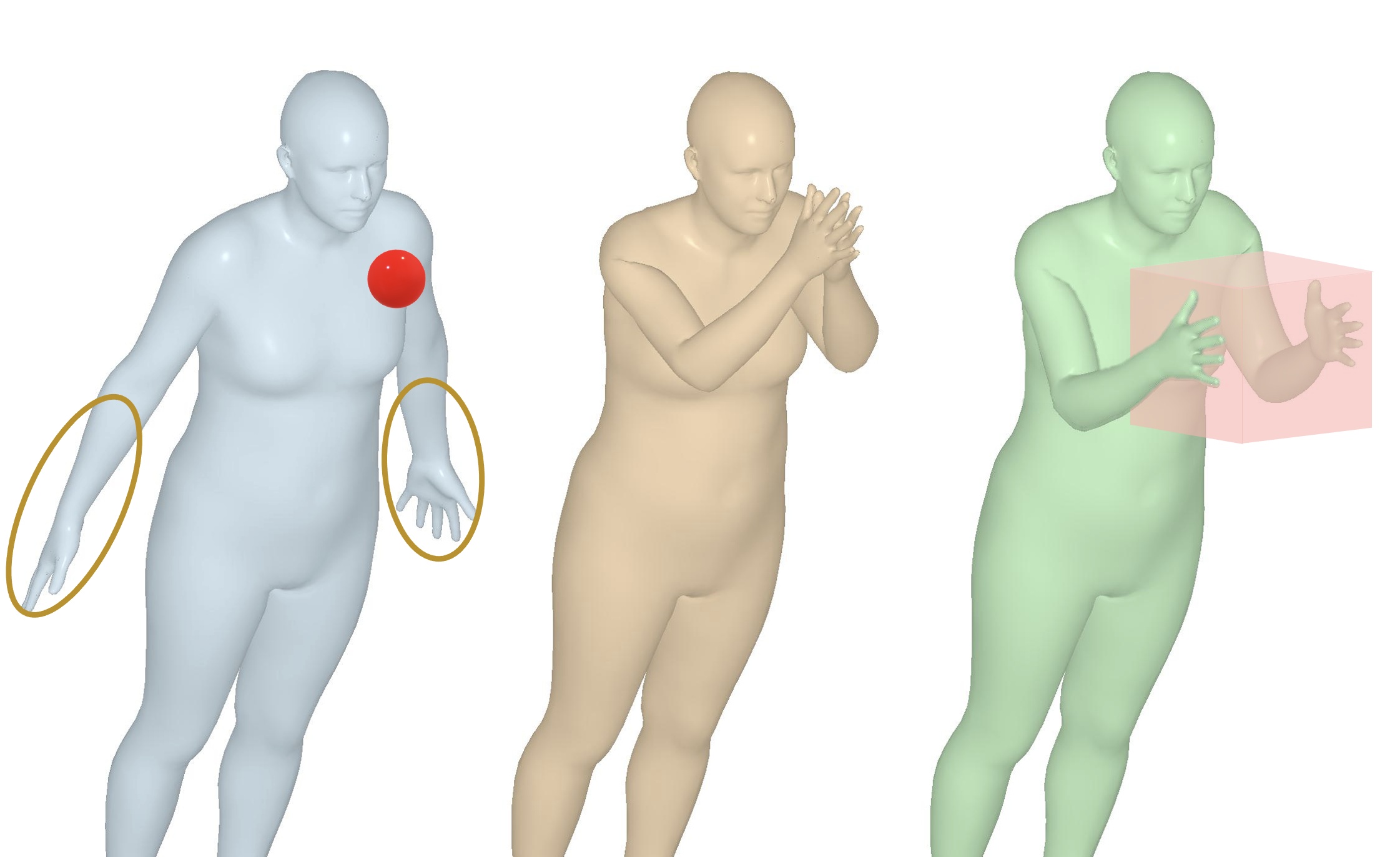}
   \caption{
       \small{{\bf Interacting with the objects.} 
       One can build a blended field (Eq.~\ref{eq:blended_field}), where non-zero component (defined at arms, gold ovals) induces the motion, while zero component impedes it. Starting from the same posture ({\it blue body on the left}), the motion towards the target ({\it red dot in front of the face}) is computed without any constraints ({\it gold body in the middle}) and with zeroing out the field inside the red box ({\it green body on the right}). 
       The blending with the area empty from the field successfully prevents penetrations inside the area. Without constraints, both hands reach the target point.
}}
\vspace{-5mm}
    \label{fig:box_field}
\end{figure}

Another potential application of such customized integration is the interaction with objects. The subject must be able to move in the field, while not going through  objects. This can be achieved by modeling the space inside the object as a region of zero field. When blending during integration, the moving points get stuck in the non-moving area, preventing further penetration. 

We illustrate this in the example in Fig.~\ref{fig:box_field}. The target point for both hands is located in front of the face (red dot). As in the previous experiment, the field exists only for selected areas of the arms (gold ovals). During integration, the local fields of the hands are blended with the zero field of the box, and the hands stop at the surface of the box (green body on the right). If no constraints are imposed on the field (gold body in the middle), both hands successfully reach the target point.

\subsection{Ablation Study}

\parag{Picking the Initial Body Posture.}
\label{sec:pick}


\begin{table}[t]
    \begin{small}
    \begin{center}
      \scalebox{1.0}{
    \begin{tabular}{l|c|c|c}
      method & MPJPE & P-MPJPE & Accel.Err \\
      \hline
        HMR2.0~\cite{Goel23a}  & 82.0 & 52.7 & 16.1  \\
        \hline
        {\it Jitter}  & {\bf 81.6} & {\bf 52.6} & 15.9   \\
        {\it Key Poses} & 82.9 & 53.6 & 16.4  \\
        {\it Successive Frames} & 82.3 & 53.2 & {\bf 9.4} \\
    \hline
    \end{tabular}}
    \end{center}
    \end{small}
    \vspace{-2mm}
    \caption{\small {\bf Picking the initial body posture.}
    All variations of \Ours{} produce 0.0\% collision rate.
    The {\it Jitter} strategy improves position error, while {\it Successive Frames} variant better preserves the overall motion smoothness. Experiments are performed on 3DPW-test set~\cite{Marcard18}.
    }
    \vspace{-5mm}
    \label{table:abl_start}
    \end{table}

Recall from Section~\ref{sec:eliminate} that, when solving our ODE, we took the initial body position to be the corrected one in the previous frame. 
We refer to this strategy as {\it Successive Frames}. This works well for video sequences but there are cases where this would be impractical, for example when dealing with single images. We have therefore explored two alternatives.
\begin{itemize}

\item {\it Jitter.} Having localized the self-intersections, we can determine what part of the body and group of limbs are responsible for them. A straightforward solution is then to randomly jitter these parameters until a body without self-intersection is obtained. This method is simple yet proves to be highly effective in most cases.

\item {\it Keyposes.}  We first precompute a dictionary of poses with no self-intersections. To this end, we subsample the AMASS~\cite{Mahmood19} pose dataset  and cluster the poses using K-means with the keypoint distance as metric. We then take the closest neighbors to the $K$ cluster centers as keyposes. Finally, given a body we wish to correct, we take the closest keypose to be our starting point. We found $K = 128$ to be sufficient to find a suitable keypose shape in most cases. 

\end{itemize} 

In Table~\ref{table:abl_start}, we compare these different ways to initialize the \Ours{} process on the same video sequences of 3DPW-test set~\cite{Marcard18}. {\it Successive Frames} denotes the approach of Section~\ref{sec:eliminate}, which is clearly best at preserving a smooth natural motion. However, the {\it Jitter} technique yields slightly more accurate pose estimates, even surpassing those of the baselines as reported in Table~\ref{tab:col_penalizer}. The {\it Keyposes} approach performs worse than the others because it requires integration from the relatively distant body shapes. Yet, there may be some rare cases when it is the only usable, for example, when there is no neighboring frame without self-intersections and jitter in the parametric space does not yield a valid body shape.

\parag{Choosing Optimal Sampling.}
\label{sec:sampling}


\begin{figure}[t!]
 \centering
  \includegraphics[width=\linewidth]{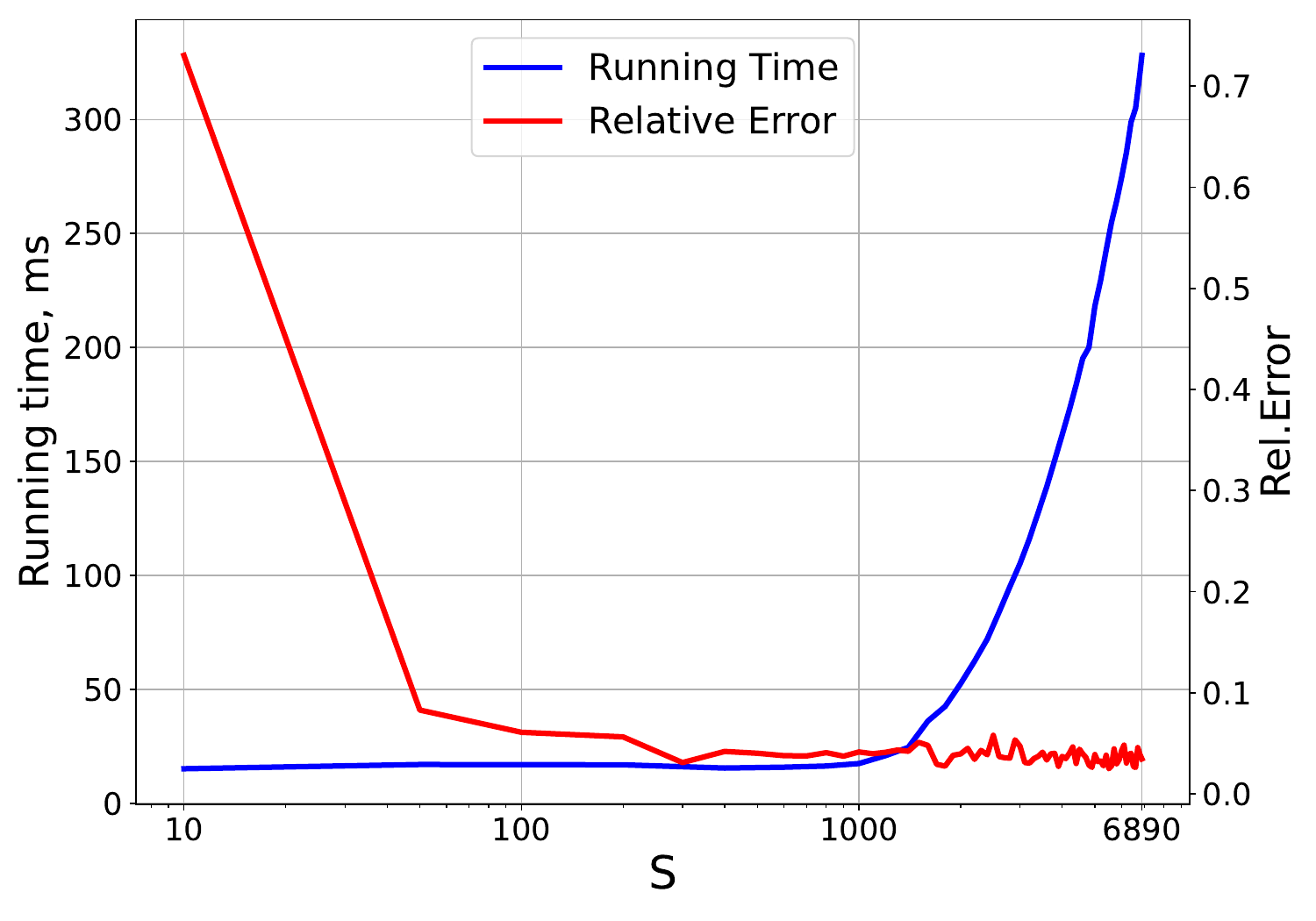}
  \vspace{-7mm}
   \caption{
       \small{{\bf Ablation on Points Sampling.}
       The Jacobian computation injects a tradeoff between the accuracy and the running time. Using too many points $S$ for the evaluation slows down the calculations, while less do not provide enough accuracy. We choose $S=1000$ as the best tradeoff.
}}
\vspace{-5mm}
    \label{fig:ablation}
\end{figure}

As discussed above, one step of \Ours{} includes the Jacobian computation with the following solving the over-constrained system of equations. It brings in a natural tradeoff between time and accuracy, since the more points we sample, the more time is spent on the Jacobian computation, but the more accurate the result is. We explore the effect of the number of points $S$ used to sample from the body shape following Eq.~\ref{eq:lsq}. 

We show the results in Fig.~\ref{fig:ablation}. The running time is the time spent only on the Jacobian computation, as all other steps are negligible. 
To estimate the accuracy of the \Ours{} step, we develop a simple yet effective metric. We sample an SMPL vector $\Theta_0$ and add a random noise $\delta\Theta$ of the known amount (we use $\|\delta\Theta\| = 10^{-2}$) to obtain the second posture $\Theta_1=\Theta_0 + \delta\Theta$. For a pair of bodies and the sampling number $S$, one can compute the distance between the bodies in the coordinate space $\bm{f} \in \mathbb{R}^{3S}$. As the linear approximation of the SMPL transformation holds for such a small $\delta\Theta$, solving $\hat{\delta\Theta} = \bJ^{\dagger}\bm{f}$ must give a solution that is close to the ``ground-truth'' $\delta\Theta$. A more elaborate reasoning behind the linear assumption we make is discussed in the supplementary material.
We define the relative error as 
\begin{equation}
    RE(\delta\Theta) = \frac{\|\hat{\delta\Theta} - \delta\Theta\|}{\|\delta\Theta\|} \; .  
    \label{eq:re}  
\end{equation}
 We choose $S=1000$ as a good tradeoff between the accuracy and the time and use it in all our experiments. 


\section{Conclusion}

We have presented an approach to reliably eliminating self-intersection from body shape and pose estimation by solving an ODE while imposing a body shape prior. Unlike methods that rely on minimizing a loss function, ours guarantees the complete disappearance of {\it all} self-intersections. Furthermore, it is differentiable, which means that it can be integrated into a deep learning training pipeline. 
We have shown how to exploit the differentiability of \Ours{} to fine-tune networks and improve their performance while decreasing the amount of self-intersections they produce.
Last but not least, we have demonstrated how our \Ours{} strategy can be applied to practically any customized motion field, for example, enabling body interaction with the environment.

In future work, we will extend \Ours{} to motion generation and exploit existing motion priors.

{
    \small
    \bibliographystyle{ieeenat_fullname}
    \bibliography{string,learning,vision,graphics,misc}
}

\renewcommand\thefigure{\thesection.\arabic{figure}} 
\renewcommand\thetable{\thesection.\arabic{table}} 
\renewcommand\thesection{\thesection.\arabic{section}} 
\renewcommand\theequation{\thesection.\arabic{equation}} 

\clearpage
\maketitlesupplementary

\appendix

The followong supplementary material is organized as follows. In Sec.~\ref{app:main_exps}, we provide more qualitative illustrations on how \Ours{} works. In Sec.~\ref{app:ablations}, we discuss the technical details of the ablation study, discussed in the main paper. In Sec.~\ref{app:custom_field}, we provide implementaiton details on the customized motion field and discuss the comparison between such simple field with the one induced by the neural network.

\section{Main Experiments}
\label{app:main_exps}


\begin{figure*}[t!]
 \centering
  \includegraphics[width=\linewidth]{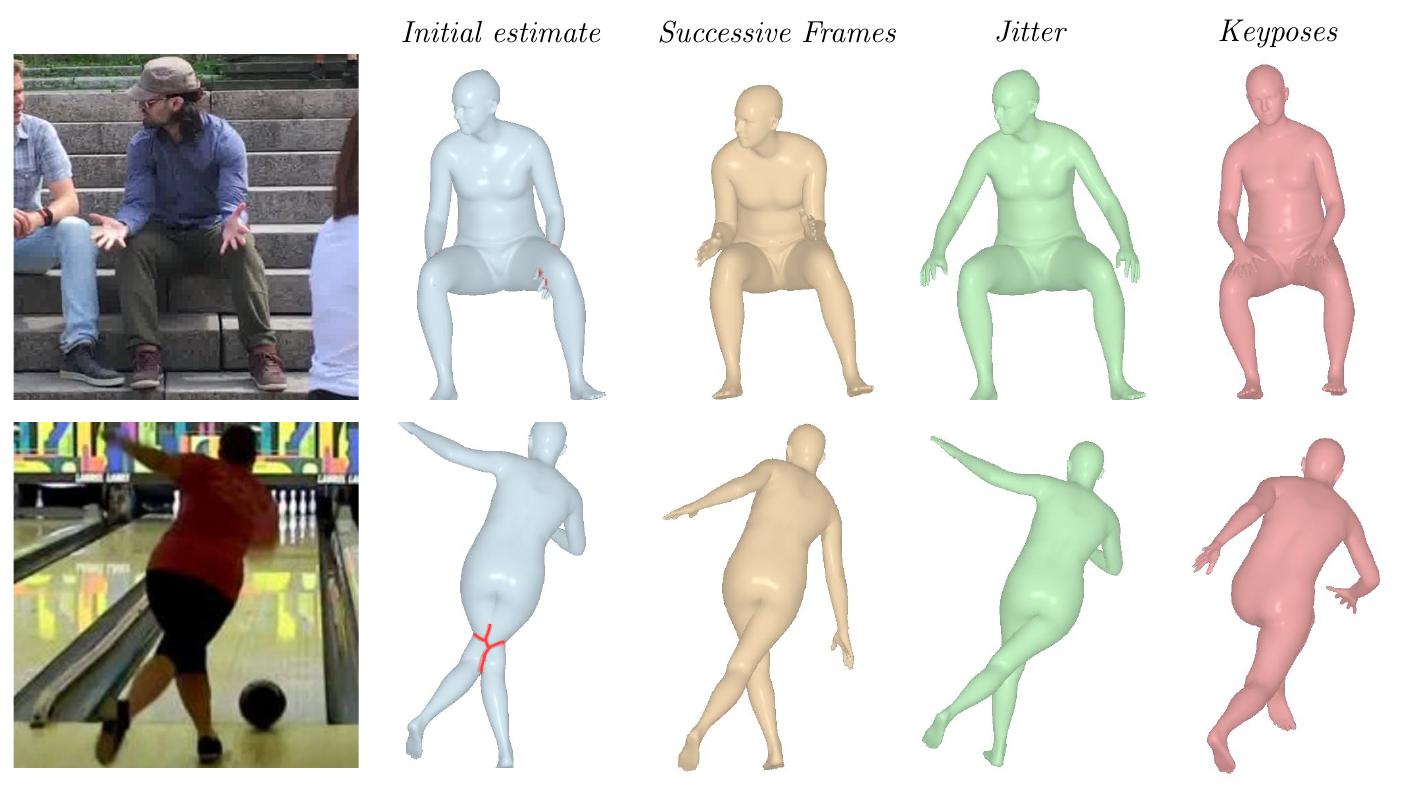}
   \caption{
       \small{{\bf Picking the Initial Body Posture.}
       We demonstrate the examples of the initial body postures $\Theta_0$ to the integration, given {\color{RoyalBlue} the initial estimate that has self-intersections}. We compare three strategies for our \Ours{} method, {\color{YellowOrange} Successive Frames}, {\color{ForestGreen} Jitter} and {\color{red} Keyposes}. None of them has self-penetrations, however, they are different in terms of the distance from the initial estimate. 
       }
}
    \label{fig:starting_bodies}
\end{figure*}

In Sec.~\ref{sec:pick} in the main paper, we discuss different strategies to obtain the initial body posture $\Theta_0$ for the integration. In Fig.~\ref{fig:starting_bodies}, we demonstrate qualitative comparison between initial bodies obtained with different strategies. {\it Successive Frames} produces a reasonable estimate, since the pose in neighboring frames is very similar. 
If {\it Jitter} manages to find an estimate without self-intersections, then it is also decent, since only a few body parameters are changed, hence, most of the non-intersected body parts are not affected at all. 
The {\it Keyposes}-produced neighbor is more distant from the given estimate, though, in some circumstances, it can be an only option. 

We have {\it not} found the choice of the starting pose for {\it Jitter} to be critical. For {\it Keyposes}, the sensitivity is larger because the key poses have been chosen to be diverse. Hence, usually only one pose among all yields a good starting estimate. It should be noted that increasing the number of key poses helps only marginally, while making the neighbor search more costly.

In video \texttt{CLOAF.mp4}, we compare HMR2.0 with \Ours{} used as post-processing on one video sequence of 3DPW-test set. As discussed in Sec.~\ref{sec:eliminate}, \Ours{} can be used for efficient removal of self-intersections. The inverse operation of Eq.~\ref{eq:smplOde} effectively averages the velocities of all points on the limb producing the realistic motion, even when limbs touch each other and points in contact have the same velocities, as shown in the video.

\section{Ablations and Technical details}
\label{app:ablations}

\parag{Optimal sampling.}

In Ablations in Sec.~\ref{sec:sampling}, we speculate that the linear approximation of the SMPL transformation holds for small displacements $\delta\Theta$. Here we verify this assumption. As in the main paper, we denote $\bm{f}$ to be the distance between two SMPL bodies $\Theta_0$ and $\Theta_1 = \Theta_0 + \delta\Theta$, computed in the coordinate space:
\begin{equation}
    \bm{f}(\delta\Theta) = \bX(\Theta_0 + \delta\Theta) - \bX(\Theta_0),
\end{equation}
where the $\delta\Theta$ is the random noise of the magnitude $\|\delta\Theta\|$. The forward SMPL transformation $\bX(\Theta)$ is computed with Eq.~\ref{eq:x_smpl}, sampling all vertices in the SMPL mesh $S=6890$, hence, $\bm{f} \in \mathbb{R}^{3S}$.  


\begin{figure}[t!]
 \centering
  \includegraphics[width=\linewidth]{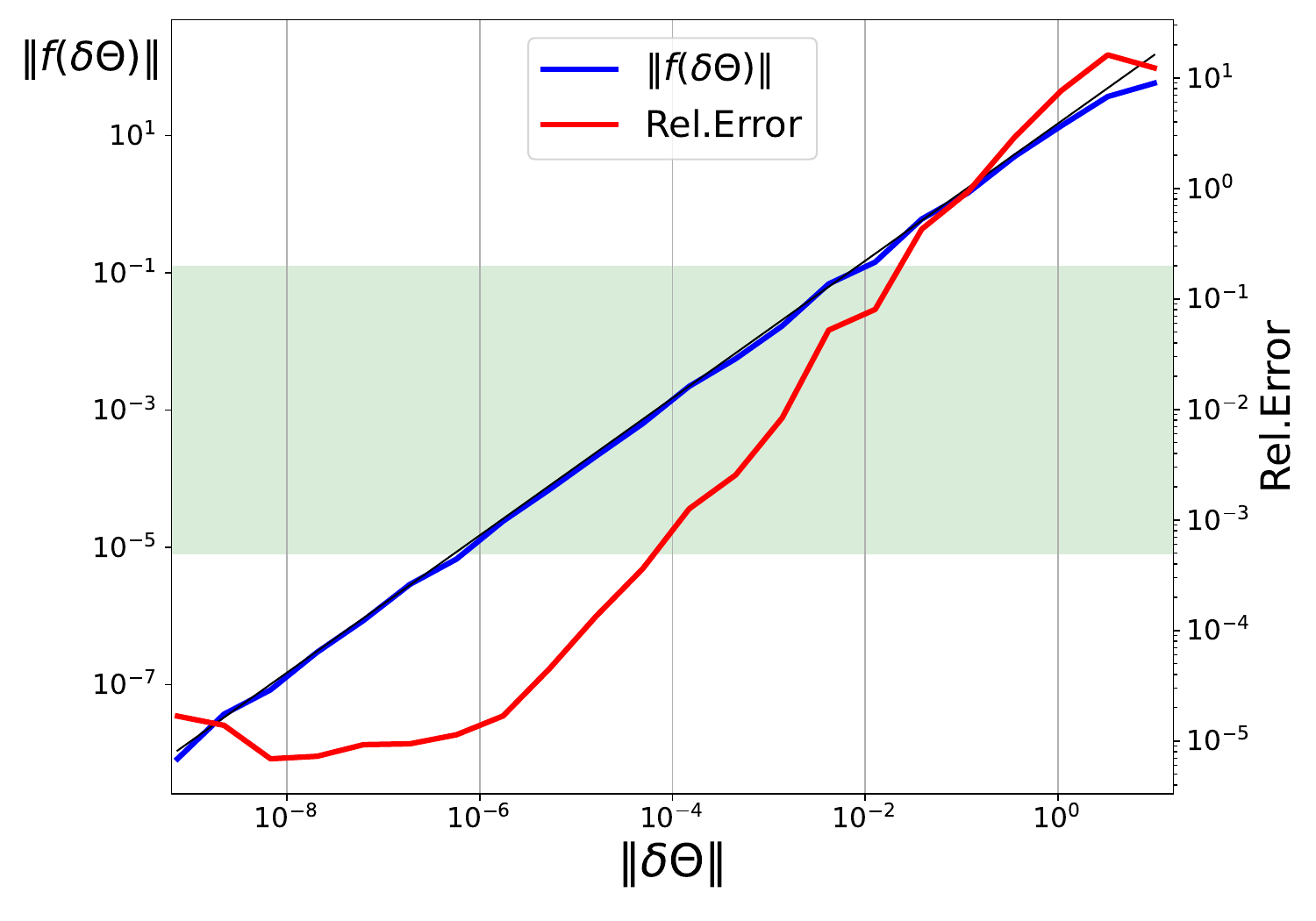}
   \caption{
       \small{{\bf SMPL linearity and the Relative Error.}
       SMPL forward transformation can be seen as linear in the wide range of $\delta\Theta$ deformations. It includes the target area (shown in green) of values that the trained network produces. 
       The Relative Error $RE$ is low in the area as well, which proves our method to be accurate.
}}
    \label{fig:linear_smpl}
\end{figure}

We provide Fig.~\ref{fig:linear_smpl} for an illustration. 
We measure the distance $\|\bm{f}(\delta\Theta)\|$ ({\it left axis}), averaged over all points, with respect to $\|\delta\Theta\|$ that varies from $10^{-9}$ to $10^{1}$. Additionally, we compute the relative error $RE$ following Eq.~\ref{eq:re} ({\it right axis}).
Every number is averaged across 10 restarts of $\delta\Theta$. The black line in the background shows the linear trend. The green area marks the range of values of the motion fields that the pretrained network $\bm{f}_{\omega}$ motion gives in our experiments. 
As assumed, for the entire range of displacements, excluding too large values $\|\delta\Theta\| > 10^{-1}$, the linear approximation holds. For larger values of $\|\delta\Theta\|$, the regime becomes non-linear, and approximation is not valid anymore, however, in our experiments, we never observe such large displacements. 
In the wide range of $\|\delta\Theta\|$ magnitudes, including the target green area, the relative error $RE$ is low ($RE\lesssim10^{-1}$). Hence, our inverse procedure is accurate. The ablation experiment in Fig.~\ref{fig:ablation} in the main paper is done for $\|\delta\Theta\|=10^{-2}$ that corresponds to the upper bound of the target regime (green area), as illustrated in Fig.~\ref{fig:linear_smpl}.

Note that all computations with the inverse Jacobian for Fig.~\ref{fig:linear_smpl} (as for all experiments of the main paper) are done in double precision, since SMPL transformation is very sensitive to single-precision roundings.
The numbers for relative error $RE$ (Eq.~\ref{eq:re}) and timing in Fig.~\ref{fig:ablation} are done by averaging multiple runs; the number of restarts is 10 for both metrics.


\begin{figure*}[t!]
 \centering
  \includegraphics[width=0.9\linewidth]{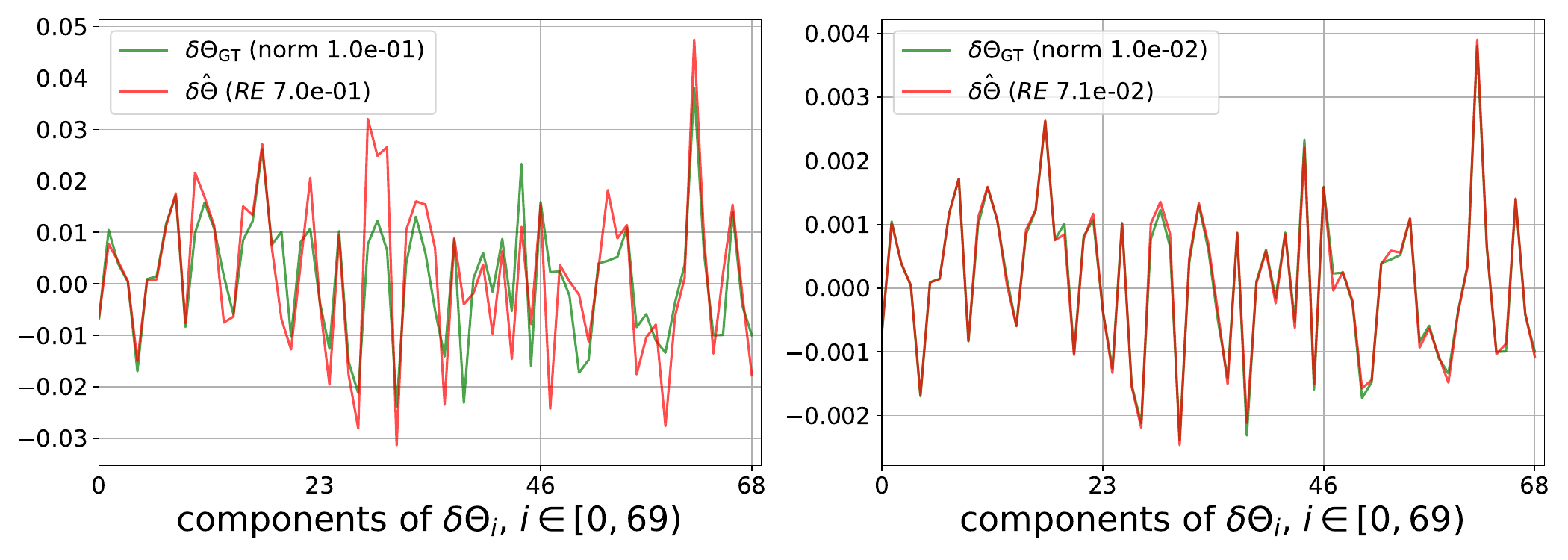}
   \caption{
       \small{{\bf Reconstruction of $\delta\Theta$ at different magnitudes.}
        We randomly sample the vector $\delta\Theta_{\text{GT}}$ and vary only its magnitude to be $10^{-1}$ and $10^{-2}$ (green). It brings reconstructions $\hat{\delta\Theta}$ of different quality, $RE = 7.0\cdot10^{-1}$ and $RE = 7.1\cdot10^{-2}$, respectively (red). The values $RE \lesssim 10^{-1}$ can be seen as an indicator of a proper reconstruction.
       }
}
    \label{fig:diff_re}
\end{figure*}

To illustrate what $RE~\lesssim 10^{-1}$ looks like, we provide Fig.~\ref{fig:diff_re}. We sample ``ground-truth'' vector $\delta\Theta_{\text{GT}}$ once (values in green) and only vary its magnitude to be $10^{-1}$ ({\it left}) and $10^{-2}$ ({\it right}). Using our inverse procedure, we reconstruct $\hat{\delta\Theta}$ (values in red). The estimate on the left is much less accurate than the one on the right, which is reflected in the relative error $RE$ ($7.0\cdot10^{-1}$ and $7.1\cdot10^{-2}$, respectively). We see that the values $RE \lesssim 10^{-1}$ can be seen as an indicator of a proper reconstruction.

\parag{Linear interpolation.}

Precise integration in Eq.~\ref{eq:smplOde} requires an estimate of $\Theta (t)$. During training, we use the approximation $\tilde{\Theta}$ to skip an inversion step and to stop error accumulation. This could yield a self-intersecting pose. But this only occurs at training time.  At inference time, we compute the entire pose sequence reusing the previous estimates. They are not self-intersecting by construction because we start from a non-penetrated body.

It might be assumed that the method works only when linear interpolant poses are not in self-intersection. However, the supplementary video \texttt{CLOAF.mp4} shows this not to be the case. When the person stands up, the hands move accordingly, even though the linear interpolation is entirely inside the body and should not produce any motion at all.

\parag{Technical Details.}

When dealing with ODEs, efficient integration is a key. To this end, we found that {\it all} techniques described is Sec.~\ref{sec:method} are crucial. Excluding one of these either prevents convergence or makes it too slow. Here we will note a few examples. 
Without the $\tilde{\Theta}$ approximation, it would take minutes per training sample, instead of less than a second. 
As for the integration step, it must be small, which is specifically enforced by the solver (Sec.~\ref{sec:method}) and the optimal sampling (Sec.~\ref{sec:sampling}).
Too long $\Delta t$ time intervals decrease stability of the training and the network does not converge to a reasonable solution. During training in all our experiments, we sample consecutive poses, hence, $\Delta t_{max} = 1$.

\section{Custom Field}
\label{app:custom_field}

In Sec.~\ref{sec:custom_field}, we propose a structure for the simplified motion field that comprises the direction from the selected region towards the target. For the sake of reproducibility, we provide here an exact formulation of such field (only its non-zero component):
\begin{equation}
    \bm{f}(\x) = F \frac{\x_T - \x}{\|\x_T - \x\|_2 + \epsilon},
     \label{eq:nonzero_field}
\end{equation}
where $\x_T$ is the target point, $\x$ is the point in the selected region, $F=10^{-3}$ is the magnitude of the field, and $\epsilon = 10^{-6}$ is a small regularization for stability. In other words, the field is represented by a vector of a fixed magnitude pointing from the selected region towards the target.

When the non-zero field described by Eq.~\ref{eq:nonzero_field} is defined, it is blended with the zero field $\bm{0}$, as described in Sec.~\ref{sec:custom_field} in the main paper. The inner and outer regions $r_{\text{in}}$ and $r_{\text{out}}$ are 10 and 30 {\it mm}, respectively. The larger values make the field less precise, while the smaller values make it more localized, hence, less smoother, which complexifies the integration.   

In the supplementary video \texttt{grab\_the\_box.mp4}, we demonstrate the interaction with objects discussed in Sec.~\ref{sec:custom_field} and depicted in Fig.~\ref{fig:box_field} in the main paper. The video shows the integration process for two fields, with and without constraints. 

\parag{Simple Field vs. NN.}

In the main experiments we exploit a pre-trained neural network to induce the motion field, while later we demonstrate that the simpler field can be used to produce customized motions. The natural question arises: why not to use the simple field for the main experiments without any neural networks at all? The answer is that the neural network is more flexible and can produce more complex motions than the target-driven field, as in Eq.~\ref{eq:nonzero_field}. As discussed in Sec.~\ref{sec:method}, the network learns to produce an interpolation in the parametric space, while moving in the coordinate space. The integration of the simple field can be seen as an interpolation in the coordinate space, while optimizing a very simple energy function 
$E = \|\x(\Theta) - \x(\Theta_1)\|$
, where $\Theta_1$ is the target pose, with infinitesimal steps $\delta\Theta$. 
Such a simple field easily gets stuck in local optima, preventing further improvement.

Yet, as shown in Sec~\ref{sec:custom_field}, the simple field approach allows for customization of motion, a task that is not as easily achieved with a neural network.

\end{document}